\RequirePackage{lineno}
\documentclass[10pt]{IEEEtran}
\usepackage{amssymb}

\usepackage{bm}
\usepackage{epsfig}
\usepackage{graphicx}
\usepackage{subfigure}
\usepackage{float}
\usepackage{cite}
\usepackage{url}
\usepackage{color}
\usepackage{balance}
\usepackage{mdwlist}
\usepackage{multirow}
\usepackage{threeparttable}

\usepackage{amsmath}
\usepackage{stmaryrd}
\usepackage{booktabs}
\usepackage{siunitx}
\let\labelindent\relax
\usepackage{enumitem}

\usepackage{threeparttable}
\usepackage{graphicx}
\usepackage{epsfig,amsmath,amsfonts,cite}
\newcommand{\thickhline}{\noalign{\hrule height 1.0pt}}
\DeclareMathAlphabet\mathbfcal{OMS}{cmsy}{b}{n}
\newcommand{\ten}[1]{\mathbfcal{#1}}
\newcommand{\mat}[1]{\mathbf{#1}}

\newcommand{\parNum}{d}

\newcommand{\uvec}{\mathbf{u} }

\newcommand{\zz}[1]{\textcolor{black}{#1}}

\newtheorem{definition}{Definition}

\def\sssp{\def\baselinestretch{0.88}\large\normalsize}\sssp

\ifCLASSINFOpdf
 
\else
\fi

\begin{document}

\title{Prediction of Multi-Dimensional Spatial Variation Data via Bayesian Tensor Completion}

\author{Jiali Luan and~Zheng~Zhang
\thanks{This work was supported by NSF-CCF 1763699, NSF-CCF 1817037 and a Samsung gift funding.}     
\thanks{Jiali Luan was with Department of Mathematics, University of California, Santa Barbara, CA 93106, USA (e-mail: luanjiali@hotmail.com).}
\thanks{Zheng Zhang is with Department of Electrical and Computer Engineering, University of California, Santa Barbara, CA 93106, USA (e-mail: zhengzhang@ece.ucsb.edu).}
}


\maketitle

\begin{abstract}

This paper presents a multi-dimensional computational method to predict the spatial variation data inside and across multiple dies of a wafer. This technique is based on tensor computation. A tensor is a high-dimensional generalization of a matrix or a vector. By exploiting the hidden low-rank property of a high-dimensional data array, the large amount of unknown variation testing data may be predicted from a few random measurement samples. The tensor rank, which decides the complexity of a tensor representation, is decided by an available variational Bayesian approach. Our approach is validated by a practical chip testing data set, and it can be easily generalized to characterize the process variations of multiple wafers. Our approach is more efficient than the previous virtual probe techniques in terms of memory and computational cost when handling high-dimensional chip testing data. 

\end{abstract}


\IEEEpeerreviewmaketitle

%
%
%
%


\section{Introduction}
Today's nano-scale semiconductor manufacturing is subject to significant process variations (e.g., uncertain geometric and material parameters caused by imperfect lithography, chemical-mechanical polishing and other steps)~\cite{variation2008}. These process variations can propagate to circuit and system levels, and cause remarkable performance uncertainties and yield degradation. Therefore, extensive numerical modeling, simulation and optimization techniques have been developed in the past decades to predict and control performance uncertainties of analog, digital and mixed-signal design~\cite{Tarek_DAC:10,Wenjian:2009,SingheeR09,xli2010,zzhang:tcad2013,zhang2017big,Wang:2004,li2004robust,zzhang:huq_tcad,zzhang_cicc2014,cui2018stochastic,cui2018uncertainty}. These numerical tools typically require a given detailed statistical model (e.g., a probability density function or a set of statistical moments) of the process variations. \zz{The statistical models of process variations are typically obtained by measuring and analyzing the performance data of a huge number of testing chips. The testing data can also be used for post-silicon yield analysis and performance tuning~\cite{li2013bayesian}.}

 It is non-trivial to design and measure testing chips. Firstly, one needs to carefully design and fabricate specialized circuits (e.g., ADC or ring oscillators) to measure or monitor the variation of certain parameters (e.g, ${\rm V}_{\rm th}$)~\cite{balakrishnan2009measurement, chang2012test, balakrishnan2011simple}. For instance, a micro-processor may have hundreds of ring oscillators to monitor parametric variations, leading to large chip area overhead~\cite{bhushan2006ring}. Secondly, one usually needs to measure many testing circuits on each die in order to extract statistical distributions or to characterize intra-die/inter-die spatial correlations. Testing these circuits can consume a large amount of time. Finally, hardware measurement may also permanently damage the chips due to mechanical stress~\cite{mann2004leading}. 

Instead of measuring all circuits, virtual testing techniques aim to reduce the cost by measuring only a small number of samples. A representative example is the ``Virtual Probe" technique~\cite{zhang2011virtual,zhang2010bayesian} and its variants~\cite{zhang2010multi,chang2011test}, which employ compressed sensing~\cite{donoho2006compressed} to estimate all performance data from a few measurement samples. In order to estimate all data on a die with an $n_1\times n_2$ array of circuits, these techniques approximate the variation data by the linear combination of $n_1n_2$ basis functions [e.g., 2-D discrete cosine transformation (DCT) bases] of spatial axises ${\rm x}$ and ${\rm y}$. When the approximation is very sparse (which is generally true in practice), the $n_1n_2$ coefficients can be estimated even if only $N< n_1n_2$ measurement samples are available. These techniques have proved to be more efficient than traditional approaches such as Kriging prediction~\cite{liu2007general} and  $k$-LSE estimation~\cite{nowroz2010thermal}. Compressed sensing is effective for processing 2-D data, but it has some shortcomings: 1) it is inefficient to exploit the structure of multi-dimensional data; 2) it involves large-scale optimization to compute all DCT coefficients. More detailed analysis will be given in Section~\ref{subsec:compare_theory} and Section~\ref{sec:result}.

This paper presents an alternative tensor approach to reduce the cost of modeling variations across multiple dies or wafers. Tensor computation~\cite{tensor:suvey} can reveal more information that cannot be captured by matrix- or vector-based computations (e.g., compressed sensing). By stacking all 2-D chip data as a multi-dimensional data array, we estimate them simultaneously with a small number of random samples. The full unknown multi-dimensional data set is characterized by several low-rank tensor factors, and the unknown tensor factors are adaptively computed by employing the recently developed variational Bayesian approach~\cite{zhao2015bayesian} with an automatic rank determination process. We demonstrate the effectiveness of our approach by a realistic data set with $717,080$ data samples describing the contact resistivity of $20$ dies, which is beyond the computational capability of Virtual Probe~\cite{zhang2011virtual,zhang2010bayesian,zhang2010multi,chang2011test}. 

\begin{figure}[t]
	\centering
		\includegraphics[width=3.3in]{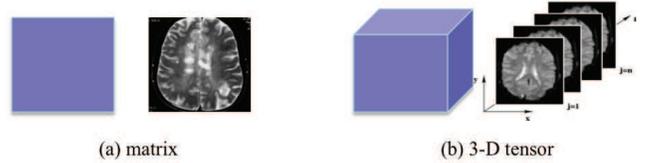} 
\caption{(a) a 2-D data array (e.g., one slice of MRI data) is a matrix, (b) a 3-D data array (e.g., multiple slices of images) is a 3rd-order tensor.}
	\label{fig:tensor}
\end{figure}


\section{Basics of Tensor}
\label{tensorref}
We first describe a few key definitions related to tensor, which are necessary to understand this manuscript. \zz{We refer the readers to~\cite{tensor:suvey} for a detailed introduction of tensor and~\cite{zhang2017tensor} for tensor computation in electronic design automation.}

\begin{definition}
A {\bf tensor} is a high-dimensional generalization of a matrix. A matrix $\mathbf{X} \in \mathbb{R}^{n_1\times n_2}$ is a $2$nd-order tensor, and its element indexed by $(i_1, i_2)$ can be denoted as $x_{i_1i_2}$. For a general $d$th-order tensor $\ten{X} \in \mathbb{R}^{n_1\times \cdots n_d}$, its element indexed by $(i_1, \cdots, i_d)$ can be denoted as $x_{i_1\cdots i_d}$. 
\end{definition}

Fig.~\ref{fig:tensor} shows a matrix and a $3$rd-order tensor, respectively. In this paper, we denote scalars by lowercase letters (e.g., $x$), vectors (tensors of order one) by boldface lowercase letters (e.g., $\mat{x}$),  matrices (tensors of order two) by boldface capital letters (e.g. $\mat{X}$), and higher-order
tensors (order three or higher) by boldface calligraphic letters (e.g., $\ten{X}$), respectively.

\begin{definition}
Given any two tensors $\ten{X}$ and $\ten{Y}$ of the same size, their {\bf inner product} is defined as
\begin{equation}
\langle \ten{X}, \ten{Y} \rangle =\sum\limits_{i_1\cdots i_d} {x_{i_1\cdots i_d} y_{i_1\cdots i_d}}.
\end{equation}
\end{definition}

\begin{definition}
Given $n$ tensors $\left \{ \ten{X}^{(m)}\right \}$ of the same size, their {\bf generalized inner product} is defined as
\begin{equation}
\langle \ten{X}^{(1)},\cdots, \ten{X}^{(n)} \rangle=\sum\limits_{i_1\cdots i_d}\prod\limits_{m=1}^n x_{i_1\cdots i_d}^{(m)}
\end{equation}
\end{definition}

\begin{figure}[t]
	\centering
		\includegraphics[width=2.0in]{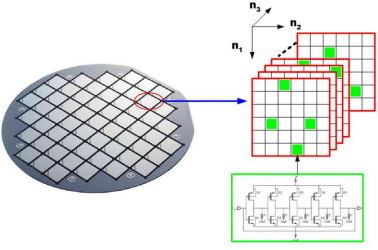} 
\caption{Tensor completion for chip data testing. Each die is a $n_1 \times n_2$ slice (i.e., a matrix) of a tensor. The small green squares represent a small number of available measurement results.}
	\label{fig:wafer_test}
\end{figure} 

Since tensors are a generalization of matrices and vectors, the above definitions of (generalized) inner product apply to matrices and vectors as well.

\begin{definition}
The {\bf Frobenius norm} of a tensor $\ten{X}$ is further defined as $|| \ten{X}||_{\rm F} =\sqrt{\langle \ten{X}, \ten{X} \rangle}$.
\end{definition}

\begin{definition}
A tensor $\ten{X} \in \mathbb{R}^{n_1 \times \cdots \times n_d}$ is {\bf rank-1} if it can be written as the outer product of $d$ vectors:
\begin{equation}
\label{eq:tensor_rank1}
\ten{X}=\uvec_1 \circ \cdots \circ \uvec_{\parNum}\; \Leftrightarrow\; x_{i_1\cdots i_{\parNum}}=\uvec_1(i_1) \cdots  \uvec_d(i_d)
\end{equation}
where $\mathbf{u}_k(i_k)$ denotes the $i_k$-th element of vector $\mathbf{u}_k \in  \mathbb{R}^{n_k}$.
\end{definition}

\section{Tensor-based Chip Testing}
\label{sec:tensor-test}

Different from the previous virtual probe techniques~\cite{zhang2011virtual,zhang2010bayesian,zhang2010multi,chang2011test} that employ compressive sensing, this section formulates the virtual testing as a tensor completion problem.

\subsection{Problem Formulation}
We consider the variations of $n_3$ dies on a wafer, and assume that each die has $n_1\times n_2$ circuits (e.g., ring oscillators) which can capture spatial correlations. Instead of measuring all $n_1 n_2 n_3$ testing circuits, we aim to estimate their performance by measuring only $N$ circuits, with $N \ll n_1n_2 n_3$. In order to achieve this goal, we first stack all dies as a 3-D data array. As shown in Fig.~\ref{fig:wafer_test}, the whole data set of die $i_1$ is a matrix $\mat{X}^{i_1}$. We can see $\mat{X}^{i_1}$ as $i_1$-th slice of a tensor $\ten{X} \in \mathbb{R}^{n_1\times n_2 \times n_3}$. Now the virtual testing problem can be formulated as tensor completion. Assume that $N$ measurement results $x_{i_1 i_2 i_3}$ are given, $(i_1,i_2,i_3)\in \Omega$, and $\Omega$ denotes the indices of measured samples. Then, we have the following problem:
\begin{equation}
\label{eq:tencomp}
{\rm Given} \; x_{i_1 i_2 i_3}\; {\rm for}\; (i_1, i_2, i_3) \in \Omega,\;  {\rm find} \; \ten{X}.
\end{equation}
This formulation can be easily extended to handle multiple wafers: one can add another index $i_4$ to indicate a specific wafer, and the whole data set $\ten{X}$ is a $4$th-order tensor.

\subsection{Low-Rank Tensor Completion}
The problem in (\ref{eq:tencomp}) has an infinite number of solutions, since we do not have any information about the un-given samples. Therefore, some constraints should be added. For instance, a $\ell _1$ regularization is used in Virtual Probe~\cite{zhang2011virtual,zhang2010bayesian} because heuristic experience shows that a 2-D DCT transform of the data on each die has very sparse coefficients. 

In this work, we estimate the unknown variation data based on a different heuristic: we find that $\ten{X}$ in the chip testing problem usually has a low-rank property in the high-dimensional space. \zz{Intuitively, this is because two reasons. Firstly, there exist strong spatial correlations. Secondly, the fabrication data samples depend on the same fabrication process, and some fabrication process have much stronger influence in causing process variations.} Similar to matrices, a low-rank tensor can be written as the sum of some rank-1 tensors:
\begin{equation}
\label{eq:tensor_rank_r}
\ten{X}=\sum\limits_{j=1}^r {\uvec_1^j \circ \cdots \circ \uvec_{\parNum}^j}.
\end{equation}
This factorization is called the CANDECOMP/PARAFAC (CP) factorization, which is one of several popular factorization formats~\cite{tensor:suvey}. Having a few samples of $\ten{X}$, we attempt to compute the factors in (\ref{eq:tensor_rank_r}) and to determine the rank $r$. Many tensor completion methods were introduced, but most approaches tend to have an inaccurate tensor rank and latent factors estimation, and eventually lead to the problem of over-fitting or weak predictive performance. 
In this paper, we choose to employ the variational Bayesian CP factorization model~\cite{zhao2015bayesian} to solve our problem. We will introduce the key ideas of variational Bayesian CP factorization in Section~\ref{sec:BayesianCP}. 
\begin{figure*}[t]
	\centering
		\includegraphics[width=4.8in]{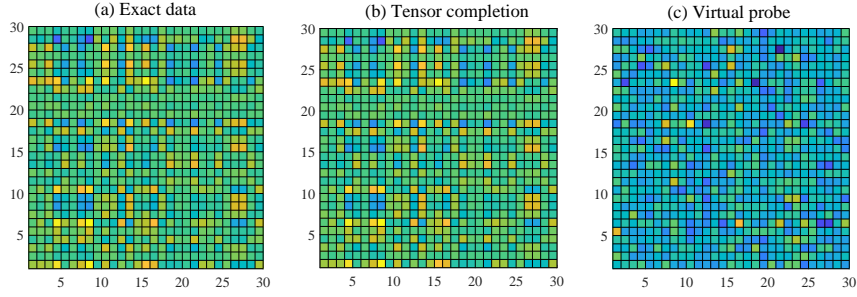} 
\caption{(a) One slice of the original tensor; (b) one slice of the result from tensor completion with $10\%$ samples; (c) one slice of the result obtained by virtual probe with $10\%$ samples.}
	\label{fig:randTensor}
\end{figure*}

\subsection{Comparison with Virtual Probe}
\label{subsec:compare_theory}
The tensor completion approach can be considered as a more flexible generalization of the virtual probe~\cite{zhang2010multi}. In order to demonstrate this, we consider the problem of approximating a $d$-variable function $f(t_1, \cdots, t_d)$, where $t_i \in [0, T_i]$ is a continuous variable for $i=1,2,\cdots, d$. 

We discretize $[0, T_i]$ into $n_i-1$ segments of length $\Delta_i= T_i/ (n_i-1)$, then the $(i_1,i_2,\cdots, i_d)$-th element of $\ten{X}$ can be regarded as the discretized value of $f(t_1, \cdots, t_d)$:
\begin{equation}
x_{i_1 \cdots i_d}=f\left( t_1=\left( i_1-1 \right) \Delta _1, \cdots, t_d= \left( i_d-1 \right)\Delta _d\right).
\end{equation}
Consequently, the low-rank tensor factorization is equivalent to approximating $f$ by some separable functions
\begin{equation}
\label{cp:func}
f(t_1, \cdots, t_d)=\sum\limits_{j=1}^r {u_1^j(t_1) u_2^j (t_2)} \cdots u_d^j(t_d),
\end{equation}
and the vector $\mat{u}_k^j$ in \eqref{eq:tensor_rank_r} includes $n_k$ discretized function values of $u_k^j(t_k)$ on the grid points $\{t_k=(i_k-1) \Delta_k \}$. In a low-rank tensor factorization, we {\it adaptively} find some {\it unknown} univariate functions $u_k^j(t_k)$ to approximate $f(t_1, \cdots, t_d)$. Once some ``good" univariate functions are found, a small number of product terms can be used to approximate $f$ (and $\ten{X}$ accordingly) with good accuracy.

The virtual probe technique~\cite{zhang2011virtual} is equivalent to approximating $f(t_1, \cdots, t_d)$ by some {\it given} and {\it fixed} basis functions:
\begin{equation}
f(t_1, \cdots, t_d)=\sum\limits_{i_1=1}^{n_1}\cdots \sum\limits_{i_d=1}^{n_d} c_{i_1\cdots i_d} {\hat{u}_1^{i_1}(t_1) \hat{u}_2^{i_2}j (t_2)} \cdots \hat{u}_d^{i_d}(t_d). \nonumber
\end{equation}
Here $\{ \hat{u}_k^{i_k} (t_k)\}$ are some {\it pre-defined} basis functions, and $\{ c_{i_1\cdots i_d}\}$ are the unknown weights. In~\cite{zhang2011virtual}, the basis functions are specifically chosen as some Fourier basis functions. This choice of basis functions normally leads to a sparse representation; however, it is not guaranteed optimal. For instance, the Fourier basis function is not a good choice to approximate a non-smooth function $f(t_1, \cdots, t_d)$. Non-smooth behaviors actually frequently appear in spatial variation modeling due to the random systematic variations~\cite{stine1997analysis}. In fact, $f(t_1, \cdots, t_d)$ is rarely smooth with respect to $t_k$ when dimension $k$ is not an actual spatial dimension (e.g., when $k$ is the additional dimension after stacking multiple 2-D dies as a 3-D array). 

Our tensor-completion chip testing approach is more flexible and often more efficient than the virtual probe technique because of the following reasons:
\begin{itemize}[leftmargin=*]
	\item In tensor completion, the univariate function $u_k^j(t_k)$ in \eqref{cp:func} is chosen {\it adaptively} and is not limited to smooth functions (e.g., cosine functions as in~\cite{zhang2011virtual}). Therefore, tensor completion is able to use ``better" univariate functions and thus fewer product terms to approximate $f(t_1,\cdots, t_d)$. For instance, we consider a random low-rank 3D data array of size $30\times 30 \times 15$. We use tensor completion to recover the whole 3D data set with $10\%$ entries, and use virtual probe to recover the data slice by slice. Fig.~\ref{fig:randTensor} shows the result of the first slice. The Virtual Probe has a huge error (and a relative error of $106\%$) for this data set due to the non-smoothness. In contrast, tensor completion recovers the non-smooth data perfectly with a relative error $2.23\times 10^{-9}$.  
	\item The virtual probe technique forms an under-determined equation with $N=\prod_{k=1}^d n_i$ columns to compute the unknown weights $\{ c_{i_1\cdots c_d} \}$. As $d$ increases, the size of this linear equation grows exponentially. Consequently, importing this huge-size equation is beyond the capability of a computer's physical memory, let alone performing numerical computation. Our tensor completion approach is a natural solution for large $d$, and it does not suffer from the curse of dimensionality.
\end{itemize}

\section{Bayesian CP Factorization}
\label{sec:BayesianCP}
We employ the variational Bayesian approach~\cite{zhao2015bayesian} to solve our problem, due to its automatic rank determination and low computational cost. The key ideas are summarized below.

Let $\ten{Y}=\ten{X}+\varepsilon$ be a noisy tensor, and the true latent tensor $\ten{X}$ is generated by a CP model:
\begin{equation}
\label{eq:ten_cp}
\ten{X}=\sum_{j=1}^{r}\mathbf{u}_1^{j}\circ\cdots\circ\mathbf{u}_d^{j}=\mathbb{T}_{cp}(\mat{U}_{1},...,\mathbf{U}_{d}).
\end{equation}
The noise term $\varepsilon$ is an i.i.d Gaussian distribution, $\varepsilon\sim\prod_{i_1,\cdots,i_{d}}\mathcal{N}(0,\tau^{-1})$. Here, we have used $\mat{u}_k^j$ to denote the factor vector of dimensionality $k$ in the $j$-th outer product; we use the matrix $\mat{U}_k=[\mat{u}_k^1,\cdots,\mat{u}_k^r]\in\mathbb{R}^{n_k\times r}$ to denote all factor vectors associated with the $k$-th dimension.

{\bf Probabilistic Model.} Suppose $\Omega$ denotes the indices of some observed entries in $\ten{Y}$, the observed tensor $\ten{Y}_{\Omega}$ is defined as
\begin{equation*}
    \ten{Y}_{\Omega}= \begin{cases}
               y_{i_1\ldots i_d} &  \text{if}\quad (i_1,\ldots, i_d)\in \Omega \\
               0              & \text{otherwise}.
           \end{cases}
\end{equation*}
We further denote $\mat{u}_{k,i_k}=[\mat{U}_k(i_k,:)]^T\in\mathbb{R}^{r\times 1}$, i.e., the transpose of the $j$th row of matrix $\mat{U}_k$. Combining the noise distribution and the CP model, we get the observation model that is factorized over the observed tensor entries
\begin{equation*}
\begin{aligned}
p(\ten{Y}_\Omega|\lbrace\mathbf{U}&_{k}\rbrace_{k=1}^{d},\tau)=\prod_{(i_1, \ldots, i_d) \in\Omega}\\
&\mathcal{N}\left (y_{i_1...i_d}|\langle\mathbf{u}_{1,i_1},...,\mathbf{u}_{d,i_d}\rangle,\tau^{-1}\right )
\end{aligned}
\end{equation*}
The selection of a latent tensor rank, $r$, has been a challenging task. Previous probabilistic models rely on a predetermined tuning parameter chosen either by cross-validations or maximum likelihood. However, the Bayesian CP factorization method~\cite{zhao2015bayesian} is able to automatically determine the tensor rank as part of the Bayesian inference process. This approach uses a set of hyper-parameters $(\boldsymbol{\lambda}=[{\lambda}_1,{\lambda}_2,...,{\lambda}_r])$ to control the rank. Each ${\lambda}_j$ controls the magnitude of the $j$-th column of each $\mat{U}_k$. Further, a sparsity-inducing prior distribution is placed over all the latent factors, given by
\begin{equation}
\label{eq:pdf_U}
p(\mathbf{U}_k|\mathbf{\boldsymbol{\lambda}})=\prod_{i_k=1}^{n_k}\mathcal{N}\left(\mathbf{u}_{k,i_k}|\mathbf{0},\mathbf{\Lambda}^{-1}\right),\forall k \in [1,d]
\end{equation}
where $\mathbf{\Lambda}$=diag$(\boldsymbol{\lambda})$ is the precision matrix. Note that the larger ${\lambda}_j$, the smaller elements in the $j$-th column of $\mat{U}_k$. The hyper-priors over $\mat{\boldsymbol{\lambda}}$ and $\tau$ are Gamma distributions, factorized over each dimensionality of the latent tensor $\ten{X}$, given by
\begin{equation}
\label{eq:pdf_lambda}
p(\boldsymbol{\lambda})=\prod_{j=1}^r\text{Ga}(\lambda_j|\mat{c}_0(j),\mat{d}_0(j)),
\end{equation}

\begin{equation}
p(\tau)=\text{Ga}(\tau|a_0,b_0),
\end{equation}
where $a_0,b_0$ (scalars) and $\mat{c}_0, \mat{d}_0$ (vectors) are selected heuristically. Together, the overall probabilistic model can be expressed as the following joint distribution 
\small
\begin{equation*}
\begin{aligned}
p(\ten{Y}_\Omega,&\mat{U}_{1},\ldots,\mat{U}_{d},\boldsymbol{\lambda},\tau)=p(\ten{Y}_\Omega|{\lbrace\mat{U}_{k}}\rbrace_{k=1}^d,\tau)\prod_{k=1}^dp(\mat{U}_{k}|\boldsymbol{\lambda})p(\boldsymbol{\lambda})p(\tau).
\end{aligned}
\end{equation*} 
\normalsize



{\bf Variational Bayesian Solver.} We need to compute the posterior distribution of all latent variables, including CP factors $\{\mat{U}_k \}$, rank-controlling hyper-parameters $\boldsymbol{\lambda}$, and noise precision $\tau$. In order to achieve this goal, the variational Bayesian method~\cite{zhao2015bayesian} seeks for a distribution function $q(\mat{U}_1,\ldots,\mat{U}_d,\boldsymbol{\lambda},\tau)$ by minimizing the KL divergence of the two distribution $p$ and $q$. By assuming that the posterior distributions of $\mathbf{U}_k$ , $\lambda$, and $\tau$ are independent, one can use \zz{expectation-maximization (EM)} steps to obtain the optimal solution in the following form:
\begin{equation}
q_k(\mathbf{U}_k)=\prod_{i_k=1}^{n_k}\mathcal{N}(\mathbf{u}_{k,i_k}|\mathbf{\widetilde{u}}_{k,i_k},\mathbf{V}_{k,i_k}),\forall k \in[1,d]
\label{eqn:factor_posterior}
\end{equation}
\begin{equation}
q_{\lambda}(\boldsymbol{\lambda})=\prod_{j=1}^{r}\text{Ga}({\lambda}_j|c(j),d(j))
\label{eqn:lambda_posterior}
\end{equation}
\begin{equation}
q_{\tau}(\tau)=\text{Ga}(\tau|a,b)
\label{eqn:tau_posterior}
\end{equation}
where the computed $c(i)$ and $d(j)$ determines the posterior mean of $\lambda_j$ and thus the magnitude of the $j$-th column of all CP factors $\{ \mat{U}_k\}_{k=1}^d$.

The predicted tensor $\ten{\widetilde{Y}}$ is calculated based on the posterior mean of the obtained latent factors:
\begin{equation}
\ten{\widetilde{Y}}=\sum_{j=1}^{r}\mat{\widetilde{u}}_1^{j}\circ\cdots\circ\mat{\widetilde{u}}_d^{j},
\end{equation}
where $\mat{\widetilde{u}}_k^{j}$ is the posterior mean of vector $\mat{u}_k^j$. The relative error of the predicted data can be measured as
\begin{equation}
\text{relative error}=||\widetilde{\ten{Y}}-\ten{Y}||_{\rm F}/||\ten{Y}||_{\rm F}.
\end{equation}

\begin{figure}[t]
	\centering
	\includegraphics[width=3.3in, height=2.2in]{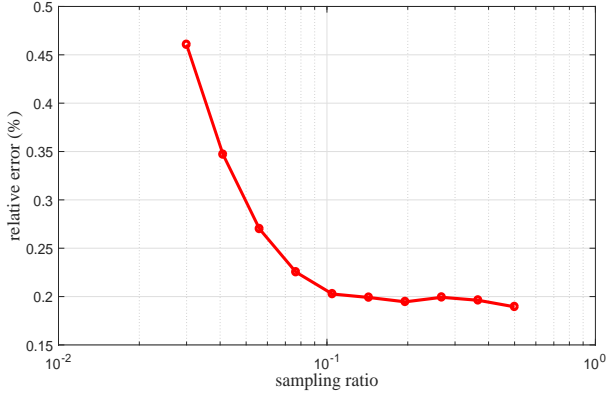} 
\caption{Relative errors of tensor completion with various sampling ratios.}
	\label{fig:Relative_Error_Plot}
\end{figure}

\section{Numerical Results}
\label{sec:result}

In order to verify our tensor completion-based chip variation data prediction method, we have implemented the algorithm in MATLAB on a Windows Desktop Workstation with 8-GB RAM and a 3.4-GHz CPU. We test our codes by a data set describing the variability of contact plug resistance in a 90nm CMOS process~\cite{balakrishnan2009measurement}. The data set has the measurement results of multiple dies, and each die has 256 x 144 = 35,854 testing circuits. We stack the data of 20 dies as a 3rd-order tensor, which has 717,080 data samples in total. In the numerical experiments, we assume that only a small number of samples (which are randomly picked with a uniform distribution) are given, and we aim to accurately estimate the whole 3-D data set based on the given measurement samples. 

\begin{figure}[t]
	\centering
	\includegraphics[width=3.3in]{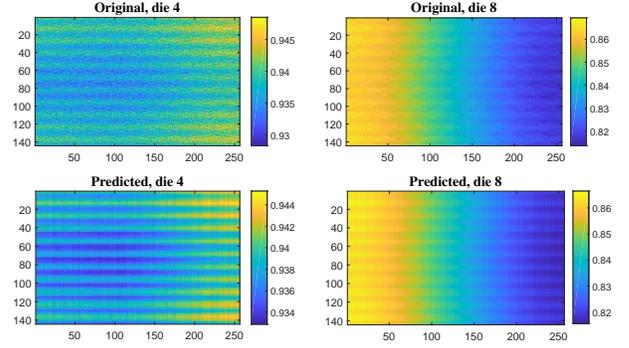} 
\caption{Top: exact spatial variation patterns of of Die 4 and Die 8. Bottom: predicted variation patterns by our approach with a $15\%$ sampling rate.}
	\label{fig:data_all}
\end{figure}

{\bf Numerical Accuracy.} In order to check the numerical accuracy of our approach on the chip data set, we fix the maximum tensor rank as 15. Then, we perform tensor completion repeatedly for different sampling ratios. The sampling ratios are chosen as 10 logarithmically spaced points between 3\% and 50\%. For each experiment, we compute the relative error of the predicted results by $||\ten{\widetilde{Y}}-\ten{Y}||_{\rm F}/{||\ten{Y}||_{\rm F}}$. As shown in Fig.~\ref{fig:Relative_Error_Plot}, our approach can predict the spatial variation data with a very small sampling ratio: the relative error decreases to around $0.2\%$ as the sampling ratio is greater than $10\%$; the relative errors are below $1\%$ for all 10 experiments.  As shown in~\cite{balakrishnan2009measurement}, the chip variation data typically has some certain patterns in the spatial domain. However, these patterns are not easy to capture, since they depend on very small variations across a die or a wafer. Our approach is capable of predicting the spatial patterns of the multiple-die data set simultaneously. We show the results of tensor completion by using a $15\%$ sampling ratio and by fixing the maximum rank as $15$. The top part of Fig.~\ref{fig:data_all} shows the exact variation patterns of two chips obtained by measuring all testing circuits. The bottom of Fig.~\ref{fig:data_all} shows the predicted variation pattern by tensor completion using only $15\%$ random measurement data. 

\begin{table} [t]
\caption{CPU Time for Virtual Probe and our method with 15\% samples.} 
\centering 
\begin{tabular}{|c |c |c |} 
\hline
Methods& Slice-By-Slice & 3-D Array  \\  \thickhline 
\\[-0.7em]
Virtual Probe & 62,947 s (17.5 h)& Out of Memory \\ \hline
Proposed Method & 325 s (5.5 min)& 67 s  \\

\hline 
\end{tabular}
\label{table:VP_Comparison}
\end{table}

{\bf Comparison with Virtual Probe.}
 We further compare our approach with the virtual probe on this realistic data set. Our data set has $144 \times 256\times20$ entries. Suppose we observe $15\%$ of the data, the Virtual Probe approach then has to solve a linear equation with 110,592 rows and 737,280 columns. Computing such a large-scale matrix and importing it into the physical memory is far beyond the capability of our desktop computer. In contrast, the Bayesian tensor completion takes only one minute to predict such a large-scale 3-D data array with an accuracy of 0.2\%, as shown in Table~\ref{table:VP_Comparison}. Since the Virtual Probe technique is unable to directly process the high-volume 3-D data, we perform another round of comparison by predicting the 3-D data array slice-by-slice. Specifically, we use Virtual Probe and tensor completion to predict the 20 individual slices of $144 \times 256$ matrices based on $15\%$ given samples. The Virtual Probe approach can work in this case, and it generates one model for each individual slice. However, Virtual Probe is extremely time-consuming: it takes 17.5 hours to predict all slices as shown in Table.~\ref{table:VP_Comparison}. In contrast, our Bayesian tensor completion finishes the prediction in only 5.5 minutes and with the similar level of accuracy. This is because that Virtual Problem has to solve a large-scale under-determined equation, whereas tensor completion only needs to compute a small number of unknown low-rank factors.

{\bf Remarks.} Our proposed approach employs the variational Bayesian method~\cite{zhao2015bayesian} to estimate the tensor rank  probabilistically. Once the algorithm converges, we can compute the expected value of each $\lambda_j$: a large $\lambda_j$ indicates very small $\mat{u}_k^j$ for all $k=1,\cdots, d$, thus the $j$-th outer product in (\ref{eq:ten_cp}) will vanish, and a tensor rank deficiency is detected. We should choose a maximum rank that is greater than the true rank; otherwise, some tensor factors cannot be captured. However, if the selected maximum rank is too large, extensive data will be required to infer the latent variables, causing higher computational cost. Table.~\ref{table:rank_prediction} has shown the predicted ranks w.r.t. different maximum ranks when the sampling ratio is fixed as $15\%$. The predicted rank remains below 20 with the relative errors around 0.2\%; the relative error decreases as the predicted rank increases to capture some small variations. However, a large maximum rank may cause over-fitting, and may overestimate the true rank.

\begin{table} [t]
\caption{Predicted tensor ranks under different maximum ranks.} 
\centering 
\begin{tabular}{c c c c} 
\hline
Maximum Rank & Predicted Rank & Relative Error  \\  \thickhline 
\\[-0.6em]
5 & 4 & 0.214\% \\
10 & 8 & 0.206\% \\
15 & 9 & 0.200\%\\
20 & 14 & 0.198\% \\
25 & 17 & 0.196\% \\ 
\hline 
\end{tabular} 
\label{table:rank_prediction}
\end{table}

\section{Conclusion}
\label{sec:conclusion}
This paper has presented a tensor framework to predict the spatial variation data of semiconductor fabrications. Our key idea is to estimate the data of multiple dies simultaneously by performing tensor completion in a higher-dimensional data space. The approach has been implemented with a recently developed variational Bayesian approach which automatically determines the tensor rank in a probabilistic way. The numerical experiments on a contact plug resistivity variation data set has shown excellent performance. High accuracy (e.g., a $0.2\%$ relative error) has been achieved with a small (e.g., $10\%$) sampling ratio. The proposed approach has also correctly predicted the spatial patterns of multiple dies simultaneously. Our proposed approach have easily handled a huge 3-D data set in one minute, whereas the Virtual Probe technique failed to work due to its huge cost of physical memory and computational resources.




\bibliographystyle{IEEEtran}
\bibliography{RefList}

\end{document}